\documentclass{article}

\usepackage[preprint]{neurips_2020}
\usepackage[T1]{fontenc}    % use 8-bit T1 fonts
\usepackage{hyperref}       % hyperlinks
\usepackage{url}            % simple URL typesetting
\usepackage{booktabs}       % professional-quality tables
\usepackage{amsfonts}       % blackboard math symbols
\usepackage{nicefrac}       % compact symbols for 1/2, etc.
\usepackage{microtype}      % microtypography

\usepackage{graphicx,subfigure}
\usepackage[cmex10]{amsmath}
\usepackage{flushend}
\usepackage{bm}
\usepackage{amssymb}
\usepackage{color}
\usepackage{dsfont}
\usepackage{amsthm}

\usepackage{algorithm}
\usepackage{algorithmicx}
\usepackage{algpseudocode}
\usepackage{epstopdf}
\usepackage{epsfig}
\newenvironment{pf}{{\noindent\it Proof}\quad}{\hfill $\square$\par}

\theoremstyle{plain}
\newtheorem{theorem}{Theorem}
\newtheorem{lemma}{Lemma}

\theoremstyle{definition}

\theoremstyle{remark}

\DeclareMathOperator*{\argmax}{arg\ max}
\DeclareMathOperator*{\argmin}{arg\ min}
\algdef{SE}[DOWHILE]{Do}{doWhile}{\algorithmicdo}[1]{\algorithmicwhile\ #1}%

\title{Unbiased Deep Reinforcement Learning: A General Training Framework for Existing and Future Algorithms}

\author{%
  Huihui~Zhang \\
  Research Institute\\
  Techman Software Co. Ltd.\\
  Chengdu, China \\
  \texttt{huihuizhang2014@gmail.com} \\
  \And
  Wu~Huang\thanks{Correspondence author is Wu huang.} \\
  Department of Computer Science \\
  Sichuan University\\
  Chengdu, China \\
  \texttt{tmezl@126.com} \\
}

\begin{document}

\maketitle

\begin{abstract}
In recent years deep neural networks have been successfully applied to the domains of reinforcement learning \cite{bengio2009learning,krizhevsky2012imagenet,hinton2006reducing}. Deep reinforcement learning \cite{mnih2015human} is reported to have the advantage of learning effective policies directly from high-dimensional sensory inputs over traditional agents. However, within the scope of the literature, there is no fundamental change or improvement on the existing training framework. Here we propose a novel training framework that is conceptually comprehensible and potentially easy to be generalized to all feasible algorithms for reinforcement learning. We employ Monte-carlo sampling to achieve raw data inputs, and train them in batch to achieve Markov decision process sequences and synchronously update the network parameters instead of experience replay. This training framework proves to optimize the unbiased approximation of loss function whose estimation exactly matches the real probability distribution data inputs follow, and thus have overwhelming advantages of sample efficiency and convergence rate over existing deep reinforcement learning after evaluating it on both discrete action spaces and continuous control problems. Besides, we propose several algorithms embedded with our new framework to deal with typical discrete and continuous scenarios. These algorithms prove to be far more efficient than their original versions under the framework of deep reinforcement learning, and provide examples for existing and future algorithms to generalize to our new framework.
\end{abstract}

\section{Introduction} \label{sec:Introduction}
% introduction from RL to DRL
While reinforcement learning (RL) agents have received great attention and been applied to multiple areas ranging from online learning and recommender engines, natural language understanding and generation \cite{tesauro1995temporal,riedmiller2009reinforcement,diuk2008object,wiering2012reinforcement,silver2017mastering}, their applicability relies heavily on the quality of handcrafted features and the observability of state spaces. Instead, DRL equiped with deep neural networks \cite{bengio2009learning,krizhevsky2012imagenet,hinton2006reducing} can provide rich representations and learn feasible policies directly from high-dimensional sensory inputs, for the goal of better performance and extensibility in RL. Currently, DRL has achieved great success in solving several control problems, utilizing deep neural networks as powerful nonlinear function approximators \cite{mnih2015human}.

% 离散 DRL
Several solutions have been proposed to stabilize the combination of online RL algorithms with deep neural networks \cite{riedmiller2005neural,schulman2015trust}, but they have the problem that the observed sequence is non-stationary and online updates are strongly correlated. Instead, Deep Q Network (DQN) \cite{mnih2013playing,mnih2015human}, an off-policy algorithm, first achieved unprecedented success the challenging domain of classic Atari 2600 games and proved to be capable of human level performance when receiving only the pixels and the game score as inputs. They stabilize the data distribution and lower correlations in the observation sequence by experience replay, while reducing correlations with the target using periodically updating Q-values towards target. Van Hasselt et al. \cite{van2016deep} then showed the universality of overestimations due to imprecise value estimates, insufficiently flexible function approximation \cite{thrun1993issues} and noise \cite{hasselt2010double,van2011insights}, and generalized the Double Q-learning algorithm to work with deep neural networks, i.e., Double deep Q-networks (DDQN). They demonstrated that reducing the overestimations due to poor policies is beneficial for more accurate value estimates. To take full advantage of action learning while holding its negative effect on the algorithm, Wang et al. \cite{wang2015dueling} adopted the dueling network so that the state-dependent action advantage function can be separated from the state value function. Under the condition of multiple similar-valued actions existing, the dueling DQN can lead to better policy and show good performance on the Atari 2600 domain. In view of sampling inefficiency, a Prioritized Experience Replay \cite{schaul2015prioritized} developed a framework for prioritizing experience with important transitions assigned higher sampling probability and replayed more frequently. Other studies tried to improve the data efficiency in DRL by replacing the state-action value function with successor representation (SR), adding feedback to observation like the uncertainty about the state-action values and exploration incentive based on the prediction error, or combining mixed techniques like model-based learning and unsupervised learning \cite{kulkarni2016deep,moerland2017efficient,stadie2015incentivizing,racaniere2017imagination,jaderberg2016reinforcement}.

% 连续控制
DQN cannot solve problems concerned with continuous domains due to the fact that it relies on on-policy or off-policy optimization to determine the action iteratively at every step. Two kinds of feasible approaches have been presented to discretize the continuous action space \cite{de2009ex,weinstein2012bandit,bucsoniu2013optimistic} or approximate the policy via parametric function in the framework of actor-critic \cite{grondman2011efficient,van2012reinforcement,degris2012model}. Nevertheless, as the dimensionality of action increases, these methods are confronted with difficulty to converge. Deep Deterministic Policy Gradient (DDPG) \cite{lillicrap2015continuous} combined the actor-critic, model-free algorithm with the deterministic policy gradient to operate over continuous action spaces. It learns an action-value function in an off-policy manner from episode sequence generated by a stochastic behavior policy and updates the deterministic target policy by gradient ascent on the value function \cite{hafez2019deep}, where the stochastic behavior policy is achieved by adding some "noise" to the deterministic policy. Besides, it uses "soft" target updates, rather than directly copying the weights, to improve the stability of learning. More recently, Dyna based on least squares
temporal difference (LSTD) and policy approximation (Dyna-LSTD-PA) \cite{zhong2019efficient} algorithm combines policy learning with the temporal difference (TD) learning to further improve the updating efficiency of the policy, and then derives a global error bound for theoretical proof.

%% 平行框架
%Different from DDPG, Asynchronous Advantage Actor-Critic (A3C) \cite{mnih2016asynchronous} employs a stochastic policy instead of deterministic target policy, but it is also in the framework of actor-critic policy gradient. Besides, A3C trains multiple agents in parallel and uses asynchronous gradient
%descent to optimize the deep network parameters. Related work on parallel architecture or asynchronous gradient training can be found in works \cite{li2011mapreduce,grounds2005parallel,tsitsiklis1994asynchronous,bertsekas1982distributed,nair2015massively,clemente2017efficient,wang2017pdp,liu2018parallel}, where each thread contains an actor equipped with a separate replay memory interacting independently with the environment, and an critic optimizes the policy/value parameters by computing the gradient of concerned algorithm loss, and then individual gradients are asynchronously sent to the global parameter for updating, and finally the global parameter periodically replace individual local parameters.

% 贡献
The contributions of this paper are fivefold. First, we organize the general loss function for DRL and propose the Unbiased Deep Reinforcement Learning (UDRL) framework. Second, we give proof to its property of uniform convergence and raise a condition for its policy improvement. Third, we analyze the cause for its efficiency and design an Enhanced UDRL as the compromise between sample efficiency and convergence rate. Fourth, several typical algorithms respectively applicable to discrete action spaces and continuous control problems are proposed based on both UDRL and Enhanced UDRL. Fifth, we analyse their computational efficiency, stability, convergence rate and sample efficiency by comparing with traditional DRL framework. %After that, we successfully apply UDRL framework to multiple parallel threads and asynchronous training \cite{mnih2016asynchronous} which trains multiple agents in parallel and uses asynchronous gradient descent. Last but not least, we seek a method that can relieve the requirement for prior knowledge of the space area.

\section{Approach}

\subsection{Related DRL}

DRL was initially applied to the scenario of discrete-action Markov decision processes (MDP) by the DQN algorithm \cite{mnih2015human}. Subsequently, its core experience replay was typically adopted by DDPG \cite{lillicrap2015continuous} to work on RL problems in continuous state and action spaces. In DRL, the agent will continually interact with the environment to achieve a sequence of observations until a terminate state arrives, which is called as an episode. During each episode, the agent will choose actions following a behavior policy to receive rewards and determine next states, which are used to train the network parameters and update the target Q-values instantly or periodically. Generally, the action-value function is represented as the discounted cumulative reward with respect to state and action, that is
\begin{align}
Q_{\pi}(s,a)=\mathbb{E}_{p^{\pi}(h|s_0,a_0)}\left[\sum_{t=0}^{\infty}\gamma^t r(s_t,a_t)|s_0=s,a_0=a\right]   ,          \label{eq:Q_value}
\end{align}
where $r(s,a)$ is the immediate reward, $s$ is the value of initial state, and $\gamma\in(0,1)$ is the discount factor for future rewards. Besides, $p^{\pi}(h|s_0,a_0)$ is the sequence probability of an episode given the initial state $s_0$ and action $a_0$, achieved by the behavior policy $\pi$.

Both DQN and DDPG are model-free, off-policy algorithms using deep neural networks. Due to the necessity of continuous control, DDPG adopts the deterministic policy gradient (DPG) technique \cite{silver2014deterministic} and actor-critic network to map states to specific actions as well as to specify the actor network and the Q network. Similar to DQN, DDPG has the current network and target network, but both equipped with individual actor-critic sub-network.

\subsection{Organized General Loss Function for DRL}

Similar to DRL, we also adopt neural networks to approximate Q-values. Then the network parameters formulate the behavior policy $\pi$ in \eqref{eq:Q_value}, and the approximation function packs the Q-value, which is originally the discounted accumulated reward of an episode sequence following MDP, into a "black box" with input $(s,a)$. No matter for discrete or continuous MDP, the network updates are originally based on the Bellman equation \cite{gattami2019reinforcement}, using one side of the Bellman equation as the target value, and we organize the general formula of the loss function for DRL as
\begin{align}
L(\omega)=\mathbb{E}_{(s,a,r,s')\sim\bm{P}}\left[(r+\gamma Q(s',\mu(s')|\omega')-Q(s,a|\omega))^2\right]   ,          \label{eq:loss_DRL}
\end{align}
where $\bm{P}$ represents the distribution probability of input $s$, $a$ is the action drawn from a behavior policy based on $s$, $r$ and $s'$ are related reward and next state achieved by interacting with the MDP environment. Given the behavior policy and MDP environment, $(a,r,s')$ can be solely determined by $s$, then $\bm{P}$ is actually the distribution of slot $(s,a,r,s')$. $\mu(s')$ is the target policy mapping $s'$ to the next action $a'$, which is normally different from the behavior policy in off-policy algorithms. $\omega$ is the parameter of neural network to be optimized that is normally different from the target network parameter $\omega'$, and $\gamma\in(0,1)$ is the discount horizon factor.

% 稳定性
An important message we can draw from \eqref{eq:loss_DRL} is that the distribution probability $\bm{P}$ of inputs don't need to follow MDP, because neural networks used to approximate \eqref{eq:Q_value} have contained all the MDP transitions. As we know, traditional DRL algorithms average the loss values over a sequence of MDP observations as the target loss for optimization to update the network parameters. However, such kind of average is inaccurate for an episode of MDP samples and thus may cause divergence or instability issues. DRL deals with the unstable problems by a biologically inspired mechanism termed experience replay \cite{mcclelland1995there,o2010play,lin1993reinforcement} that randomly samples over the history data to smooth the data distribution and alleviate data correlations in each episode. It is reasonable that with large enough size of replay buffer, the samples of experience pool can be approximatively seen as independent, which makes us doubt whether MDP are required for the sensory inputs of DRL training.

\subsection{UDRL Framework}

% UDRL
Since $\bm{P}$ in \eqref{eq:loss_DRL} does not rely on MDP, it can follow any probability distribution. The most favorable choice is the independent identically distributed (IID) sample set because it provide feasible unbiased approximation for CPU or GPU computation and facilitates computer processing.

\begin{lemma}\label{seq_iid}
If the samples of initial states $\{s_n\}_{n=1,\cdots,N}$ are IID, then the following MDP slots $(s_n,a_n,r_n,s'_n)_{n=1,\cdots,N}$ are also IID given the same behavior policy, where $N$ stands for the batch size.
\end{lemma}

\begin{theorem}\label{evalu_UDRL}
Assuming $\bm{P}$ in \eqref{eq:loss_DRL} is IID, the formula given by
\begin{align}
L_{UDRL}(\omega)=\frac{1}{N}\sum_{n=1}^N \left[(r_n+\gamma Q(s'_n,\mu(s'_n)|\omega')-Q(s_n,a_n|\omega))^2\right], \label{eq:loss_UDRL}
\end{align}
will give guarantee for the convergence of network as the number of iteration tends to infinity with properly chosen learning rate.
\end{theorem}

Based on \eqref{eq:loss_UDRL}, we propose the UDRL framework to solve the problems brought by biased RL training. An iterative update of RL algorithms is closely related to the policy at iteration $i$, which is $Q(s,a|\omega_i)\gets r+\gamma Q(s',\mu_i(s')|\omega')$, then we have
\begin{theorem}\label{impro_UDRL}
Given the condition of
\begin{align}
\frac{1}{N}\sum_{n=1}^N Q(s_n',\mu_i(s_n')|\omega')\geq\frac{1}{N}\sum_{n=1}^N Q(s_n',\mu_{i-1}(s_n')|\omega'),  \label{eq:con_Q_aver}
\end{align}
IID samples and and delayed updated target network parameterized by $\omega'$, UDRL has the property of expected policy improvement, i.e.,
\begin{align}
\mathbb{E}_{(s,a)\sim\bm{P}}\left[Q(s,a|\omega_{i+1})\right]\geq\mathbb{E}_{(s,a)\sim\bm{P}}\left[Q(s,a|\omega_{i})\right] \label{eq:exp_policy_impro}
\end{align}
\end{theorem}

The proof of Lemma~\ref{seq_iid}, Theorem~\ref{evalu_UDRL} and Theorem~\ref{impro_UDRL} can be found in Appendix. The specific processes of UDRL are given as follows. First, we uniformly randomly sample over the state space in batch to achieve independent initial state observations. After taking actions based on current states and the exploration policy, the rewards and next states can be determined, and then the target Q-values can be computed to update the network parameters. One advantage of uniform sampling is that it is the most convenient way for computers to achieve IID samples. Besides, it is also the most robust method to maximize generalization for model-free algorithms without much knowledge of the environment. Second, to explore complete information and have a stationary data distribution, the samples should cover the whole state space. In this case, we don't need memory to store large amount of past collected data to smooth the data distribution and use it to create fake unrelated samples, thus saving the costs of storing and exploiting sampled data. Instead, we train the batch composed of IID samples to update the network parameters and replace it with the batch of next iteration. The sample size for each batch may be larger or smaller than the size of mini-batch in DRL, depending on the capacity of state space, but it will be far smaller than the memory used for experience replay. By the way, of course the uniform sampling can be altered to some prioritized methods, like meta-learning \cite{botvinick2019reinforcement}, if more knowledge is provided. However, if the IID principle is violated, divergence or instability issues may also arise.

\subsection{Applications of UDRL to Discrete State-action Space}

% UDQN
When it comes to specific algorithms, we propose the Unbiased Deep Q Network (UDQN) to apply to discrete state-action space. Referring to DQN, we choose the action that maximizes the target value as the result of target policy $\mu(\cdot)$ in \eqref{eq:loss_UDRL}. Then the loss function of UDQN that need to be minimized can be given by
\begin{align}
L_{UDQN}(\omega_i)=\frac{1}{N}\sum_{n=1}^N\left[(r_{i,n}+\gamma\max_{a'}Q(s'_{i,n},a'|\omega_{i-1})-Q(s_{i,n},a_{i,n}|\omega_i))^2\right]   , \label{eq:loss_UDQN}
\end{align}
where $\omega_i$ is the parameter of neural network at iteration $i$ which parameterizes the action-value (Q-value) function as $Q(\cdot,\cdot|\omega_i)$. To some extent, $\omega_i$ is mapped to $\pi_i$, and any changes to the the behavior policy at iteration $i-1$ will lead to different updates of $\omega_i$. Accordingly, the agent's slots $(s_{i,n},a_{i,n},r_{i,n},s'_{i,n})_{n=1,\cdots,N}$ are IID samples in $i$-th batch.

% 伪代码
The pseudocode of UDQN is given in Algorithm \ref{code:UDQN}. Here we ignore the episodes and steps. Instead, we use a counter recording update/iteration numbers to track the training process.
\begin{algorithm}[h]
\caption{UDQN Algorithm}
\label{code:UDQN}
\begin{algorithmic}[1]
\State $\mathbf{Input}$: The batch size $N$, and the batch maximum $M$.
\State $\mathbf{Initialization}$: Initialize the network parameters $\omega\gets\omega_0$ arbitrarily.
\For{$i=1,M$}
\State Uniformly sample $S_i=(s_{i,1},s_{i,2},\cdots,s_{i,N})$ over the whole state space;
\State Choose actions $A_i=(a_{i,1},a_{i,2},\cdots,a_{i,N})$ for $S_i$ according to $\epsilon$ greedy policy;
\State Execute actions $A_i$, get next states $S'_i=(s'_{i,1},s'_{i,2},\cdots,s'_{i,N})$ and immediate rewards $R_i=(r_{i,1},r_{i,2},\cdots,r_{i,N})$;
\State Minimize the loss function shown in \eqref{eq:loss_UDQN} by gradient decent, and then update $\omega_i$;
\EndFor
\end{algorithmic}
\end{algorithm}

\subsection{Applications of UDRL to Continuous State-action Space}

% UDDPG
In this part, we propose the Unbiased Deep Deterministic Policy Gradient (UDDPG) to work with RL problems in continuous state and action spaces.

In continuous spaces, the greedy policy is too slow to be practically applied. Therefore, an action approximation function $\mu(s|\theta)$ is adopted to specify the deterministic policy. The current actor network is updated by maximizing the expected return, i.e., the average of action-values parameterized by the current Q network, with respect to the actor network parameter $\theta$. The formula of expected return is given by
\begin{align}
J(\theta_i)=\frac{1}{N}\sum_{n=1}^N Q(s_{i,n},a_{i,n}(\theta_i)|\omega_i),          \label{eq:a_appro}
\end{align}
where $N$ represents the sample size, $a_{i,n}(\theta_i)=\mu(s_{i,n}|\theta_i)$ and $Q(s_{i,n},a_{i,n}|\omega_i)$ is the Q-value parameterized by the critic parameter $\omega_i$ at the $n$-th sample and $i$-th iteration. The maximization of \eqref{eq:a_appro} can be achieved by gradient ascent method.

Replacing the optimal action $\mu(s'_n)$ of the target network in \eqref{eq:loss_UDRL} with the action chosen by the target actor network, which is updated partly by \eqref{eq:a_appro}, we can have the loss function to update the current UDDPG critic network, that is
\begin{align}
L(\omega_i)=\frac{1}{N}\sum_{n=1}^N\left[(r_{i,n}+\gamma Q(s_{i,n}',\mu(s_{i,n}'|\theta_i')|\omega_i')-Q(s_{i,n},a_{i,n}|\omega_i))^2\right]   ,          \label{eq:loss_UDDPG}
\end{align}
where $\mu(s|\cdot)$ is the parameterized actor network, $s_{i,n}$, $a_{i,n}$ and $s_{i,n}'$ represent the current state, current action and next state of the $n$-th sample, respectively. $\omega_i$, $\omega_i'$ and $\theta_i'$ respectively represent the parameters of current critic network, target critic network and target actor network at iteration $i$, and $(\omega_i',\theta_i')$ are soft updated by $(\omega_i,\theta_i)$, in the way of
\begin{align}
\theta_i' \leftarrow\tau\argmax_{\theta_i}J(\theta_i)+(1-\tau)\theta_i',\quad\omega_i' \leftarrow\tau\argmin_{\omega_i}L(\omega_i)+(1-\tau)\omega_i'   , \label{eq:soft_update}
\end{align}
where $\tau<1$ constrains target values to change slowly so that the stability of learning can be greatly improved. \eqref{eq:soft_update} called as "soft" target updates \cite{lillicrap2015continuous} relies on the optimization results from \eqref{eq:a_appro} and \eqref{eq:loss_UDDPG}.

% 噪声探索
To maximize the exploration during training, the noisy policy adopted in DDPG can also be employed in UDDPG, which adds independent noise samples to the actor network, i.e., the exploration policy. In that case, UDDPG will work as an off-policy algorithm due to the difference between behavior policies and target policies. Suitably chosen noise can render faster convergence rate.
Actually, UDDPG can also work on-policy.

% UDDPG与UDQN的区别
Then the distinction between UDDPG and UDQN can be summarized as: 1. the action is taken from an approximation function which is updated through the actor network instead of the Q network; 2. the "soft" target updates are adopted to keep up with the continuous action instead of periodic parameter replacement; 3. the noisy policy is employed for exploration instead of the $\epsilon$ greedy method.

% UDDPG的具体操作
When it comes to the specific operations of UDDPG, first, we uniformly sample over the continuous state space in batch to achieve independent initial state observations. Then we choose actions from an exploration policy which is combination of the current actor network parameterized by $\theta$ and the noise, and rewards and next states can be determined through interacting with the environment following the chosen actions. After that, the target Q-values are computed based on the target critic network and target actor network, whose parameters are $\omega'$ and $\theta'$, respectively.

Besides, there are three updating processes following the computation of target Q-values. First, the parameters of current actor network $\theta$ are updated by optimizing the expected return in \eqref{eq:a_appro}. Second, the parameters of current critic network $\omega$ are updated by minimizing the loss function in \eqref{eq:loss_UDDPG}. Third, the parameters of target networks $(\omega',\theta')$ are updated according to \eqref{eq:soft_update}.
After finishing the training processes, a new iteration will be launched by starting another batch of Monte-Carlo state observations. The whole process of UDDPG is organized as a pseudocode, given by Algorithm \ref{code:UDDPG}.

% 伪代码
\begin{algorithm}[h]
\caption{UDDPG Algorithm}
\label{code:UDDPG}
\begin{algorithmic}[1]
\State $\mathbf{Input}$: The batch size $N$, the batch maximum $M$, and the soft update parameter $\tau$.
\State $\mathbf{Initialization}$: Initialize the network parameters $(\omega,\theta,\omega',\theta')\gets(\omega_0,\theta_0,\omega_0',\theta_0')$ arbitrarily.
\For{$i=1,M$}
\State Uniformly sample $S_i=(s_{i,1},s_{i,2},\cdots,s_{i,N})$ over the whole state space; \label{step:MC}
\State Choose actions $A_i=(a_{i,1},a_{i,2},\cdots,a_{i,N})$ for $S_i$ according to the current actor network $\mu(S_i|\theta_i)$ added by exploration noise;  \label{step:policy_exp}
\State Execute actions $A_i$, get next states $S'_i=(s'_{i,1},s'_{i,2},\cdots,s'_{i,N})$ and immediate rewards $R_i=(r_{i,1},r_{i,2},\cdots,r_{i,N})$;
\State Maximize the expected return shown in \eqref{eq:a_appro} by gradient ascent, and then update $\theta_i$;
\State Minimize the loss function shown in \eqref{eq:loss_UDDPG} by gradient decent, and then update $\omega_i$;
\State Execute the "soft" target updates shown in \eqref{eq:soft_update} to update $\theta_i'$ and $\omega_i'$;
\EndFor
\end{algorithmic}
\end{algorithm}

\subsection{Enhanced UDRL Framework}
% 灵感来源
In this part, we set out to analyze cause for the convergence efficiency of UDRL framework. Botvinick et al. \cite{botvinick2019reinforcement} counters the sample-inefficient problem for DRL to provide a human-level learning, and describes two primary sources of sample inefficiency. It says the
first source of slowness in DRL is the requirement for incremental parameter adjustment by gradient descent, because small step-sizes in updating is necessary for stable learning. The second source is weak inductive bias, which means that DRL adopts weak initial assumption, generally less sample-efficient, to trade for a wider range of convergence. These motivate us to think about what is sacrificed to boost computational efficiency and stability for UDRL framework.

Besides the unbiased property of UDRL framework, it also has advantage over DRL in saving the costs of storing and exploiting sampled data, which consumes more memory and computation per update. And it can work well with on-policy learning algorithms. However, it has a limitation in sample efficiency because it just discards every training batch, and thus wasting the valuable samples which might be difficult to obtain. Actually, in real circumstances, large data sets may be difficult to collect due to some nuisance factors and the gap between simulations and real-world data \cite{kang2019generalization}. All of these motivate us to take full advantage of the history batch inputs and make a compromise between the convergence efficiency and sample efficiency.

% EUDRL
Therefore, we devise the enhanced version of UDRL framework, which adopts a buffer memory to store the Monte-Carlo samples per batch. Instead of training each batch directly, it randomly samples the pool of memory and trains the selected samples. Different from DRL which trains the min-batch per interaction, the enhanced version of UDRL framework will update network parameters several times before the next batch comes, which means that it will operate the randomly sampling process several times and obtain several sets of mini-batches per interaction. If the number of mini-batches per batch is exactly equal to the batch size, then each update period just consume a single sample on average, just like that DRL trains once per step.
All samples from the pool of memory are IID so it still remains unbiased. For simplicity, we call it the Enhanced UDRL (EUDRL), and pseudocodes of related algorithms EUDQN and EUDDPG are given by Algorithm \ref{code:EUDQN} and Algorithm \ref{code:EUDDPG}. For the overall setting of input constants, the memory size $D$, the maximum of updates $M$ per batch and mini-batch size $N$ of each sample from the memory are added. The agent will keep Monte-Carlo sampling and interacting with the environment to achieve IID samples until the memory is full, then it will train mini-batches of size $N$ for $M$ steps. During the training processes, the network parameters will also be updated for $M$ times. At the end of each training cycle, the oldest transition will be popped out, which contains $N'$ samples. After that, a new batch of size $N'$ will be sampled and pushed into the memory.

\begin{algorithm}[h]
\caption{EUDQN Algorithm}
\label{code:EUDQN}
\begin{algorithmic}[1]
\State $\mathbf{Input}$: The batch size $N'$, the batch maximum $M'$, the memory size $D$, the maximum of updates $M$ per batch, and the mini-batch size $N$.
\State $\mathbf{Initialization}$: Initialize the network parameters $\omega\gets\omega_0$ arbitrarily.
\For{$i=1,M$}
\State Uniformly sample $S_i=(s_{i,1},s_{i,2},\cdots,s_{i,N})$ over the whole state space;
\State Choose actions $A_i=(a_{i,1},a_{i,2},\cdots,a_{i,N})$ for $S_i$ according to $\epsilon$ greedy policy;
\State Execute actions $A_i$, get next states $S'_i=(s'_{i,1},s'_{i,2},\cdots,s'_{i,N})$ and immediate rewards $R_i=(r_{i,1},r_{i,2},\cdots,r_{i,N})$;
\State Store transition $(S_i,A_i,R_i,S'_i)$ in the memory of size $D$;
\If{memory is full}
\For{$i=1,M$}
\State Sample random mini-batches of size $N$ from the memory of size $D$;
\State Minimize the loss function shown in \eqref{eq:loss_UDQN} by gradient decent, and then update $\omega_i$;
\EndFor
\State Pop out the oldest transition $(S_{i-D+1},A_{i-D+1},R_{i-D+1},S'_{i-D+1})$.
\EndIf
\EndFor
\end{algorithmic}
\end{algorithm}

\begin{algorithm}[h]
\caption{EUDDPG Algorithm}
\label{code:EUDDPG}
\begin{algorithmic}[1]
\State $\mathbf{Input}$: The batch size $N'$, the batch maximum $M'$, the memory size $D$, the update numbers $M$ per batch, the mini-batch size $N$, and the soft update parameter $\tau$.
\State $\mathbf{Initialization}$: Initialize the network parameters $(\omega,\theta,\omega',\theta')\gets(\omega_0,\theta_0,\omega_0',\theta_0')$ arbitrarily.
\For{$i=1,M$}
\State Uniformly sample $S_i=(s_{i,1},s_{i,2},\cdots,s_{i,N})$ over the whole state space;
\State Choose actions $A_i=(a_{i,1},a_{i,2},\cdots,a_{i,N})$ for $S_i$ according to the current actor network $\mu(S_i|\theta_i)$ added by an exploration noise;
\State Execute actions $A_i$, get next states $S'_i=(s'_{i,1},s'_{i,2},\cdots,s'_{i,N})$ and immediate rewards $R_i=(r_{i,1},r_{i,2},\cdots,r_{i,N})$;
\State Store transition $(S_i,A_i,R_i,S'_i)$ in the memory of size $D$;
\If{memory is full}
\For{$i=1,M$}
\State Sample random mini-batches of size $N$ from the memory of size $D$;
\State Maximize the expected return shown in \eqref{eq:a_appro} by gradient ascent, and then update $\theta_i$;
\State Minimize the loss function shown in \eqref{eq:loss_UDDPG} by gradient decent, and then update $\omega_i$;
\State Execute the "soft" target updates shown in \eqref{eq:soft_update} to update $\theta_i'$ and $\omega_i'$;
\EndFor
\State Pop out the oldest transition $(S_{i-D+1},A_{i-D+1},R_{i-D+1},S'_{i-D+1})$.
\EndIf
\EndFor
\end{algorithmic}
\end{algorithm}

\section{Experiments} \label{Simulation}
\subsection{Maze}

\begin{figure}
  \centering
  \subfigure[]{
  \label{fig:Maze}
  \includegraphics[width=0.49\textwidth]{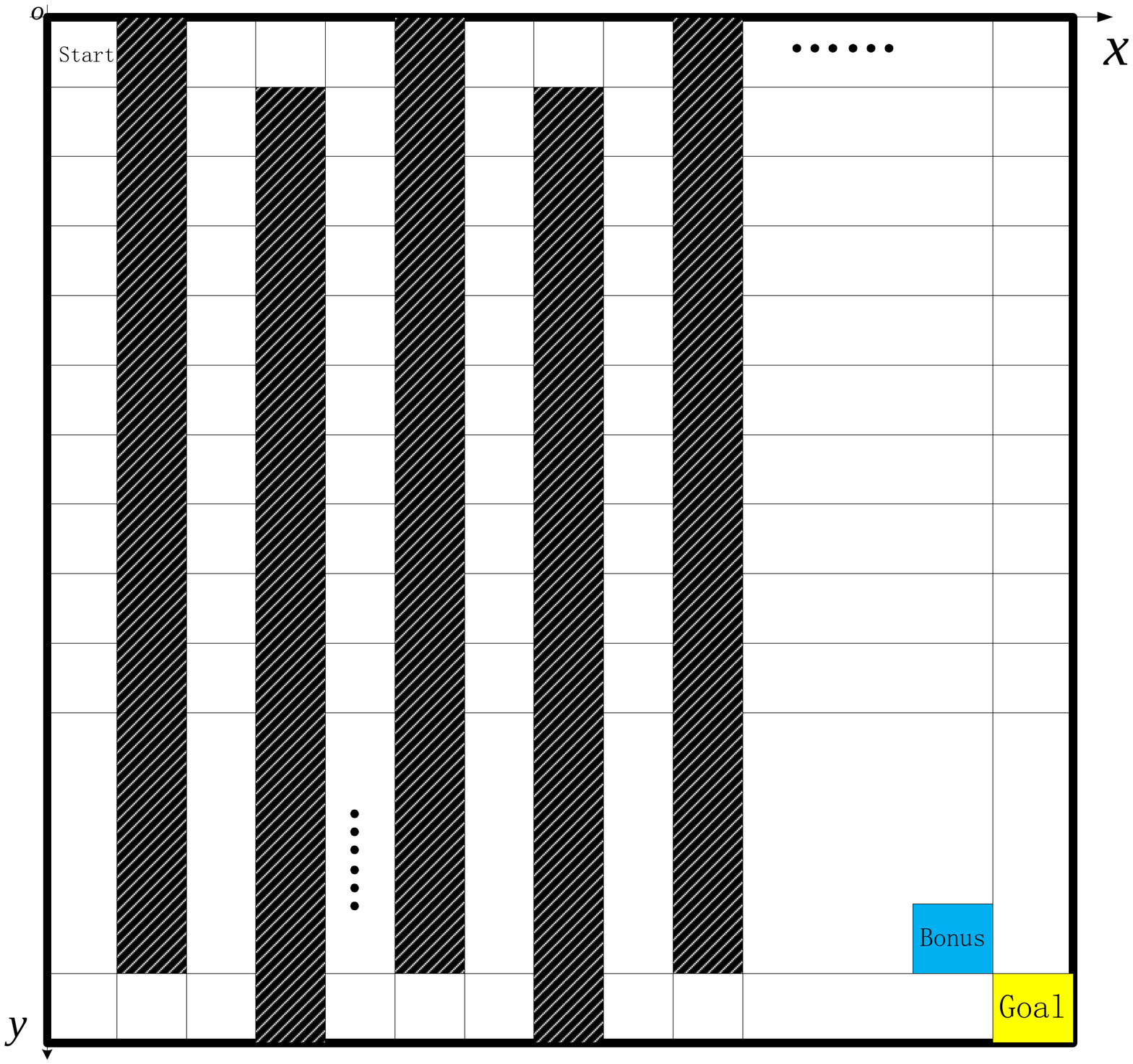}}
  \subfigure[]{
  \label{fig:Robot_arm}
  \includegraphics[width=0.49\textwidth]{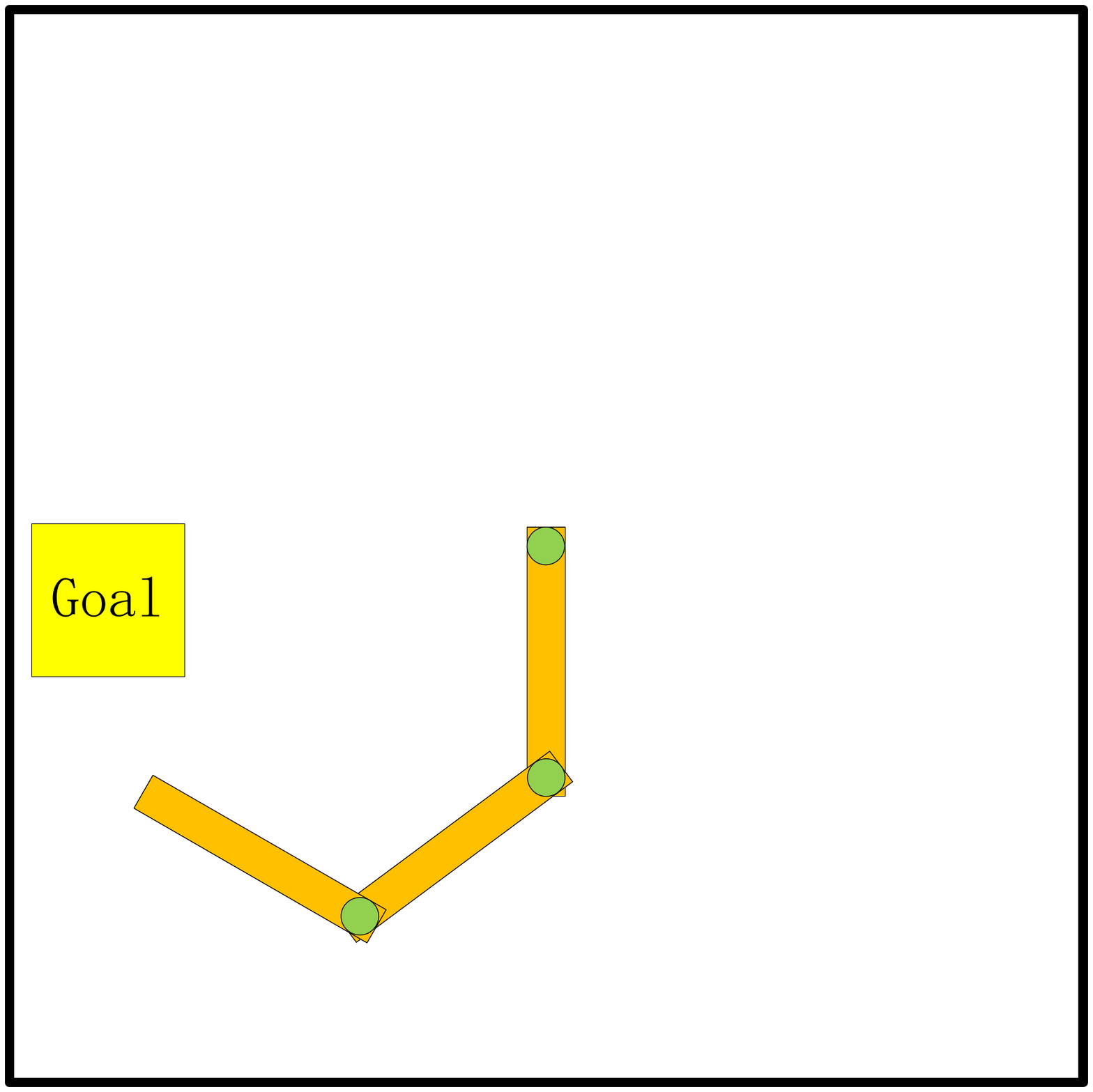}}
  \caption{\textbf{(a) The maze environment with discrete state-action spaces and endless obstacles. Due to the large state space, the whole picture is described by some of the initial states and some of the final states for simplicity. It is not difficult to conjecture the left part; (b) The robot arm environment with a grasp and move task. The finger of this robot arm is represented by the end, its task is to reach the moving goal represented by the yellow box and keep gasping for several steps.}}
\end{figure}

% Maze环境
For the performance evaluation of UDRL and EUDRL on discrete MDP, we set out to solve the maze problem with large discrete state-action spaces and endless obstacles. The environment of the maze problem is shown in Fig.~\ref{fig:Maze}. It is discrete maze environment, so each step towards four directions (up, down, left and right) takes the agent to an adjacent grid. The task of the agent is to move from the initial state labelled $start$ towards the goal avoiding the obstacles represented by the gray grids. During this process, the agent has minus reward if it touches the gray grids, which means that these areas are forbidden. Once the agent arrives at the goal, it will be rewarded $100$ scores. Besides, at the upper-left side of the goal, the state is assigned $50$ scores as a bonus reward to see whether the agent can pick up all positive rewards before reaching the goal without bumping into the obstacles. In other states represented by blank grids, the rewards are zero.

% 参数与表格 hyperparameter values； 所需先验知识
The new framework requires redefinition of the environment that can collect a batch of samples each step. To show the advantage of UDRL framework, we test both DQN and UDQN (EUDQN) algorithms for comparison, using the same hyperparameter values listed in Table~\ref{Table:par_maze}. The mere prior knowledge required for the following experiments is the domain of state space for the convenience of Monte-Carlo sampling. For a fair comparison, the DQN algorithm also exploit this prior knowledge to reset over the state space at the beginning of each episode, which can help to converge much faster.

\begin{figure}
  \centering
  \subfigure[]{
  \label{fig:UDQN-DQN-81-100}
  \includegraphics[width=0.49\textwidth]{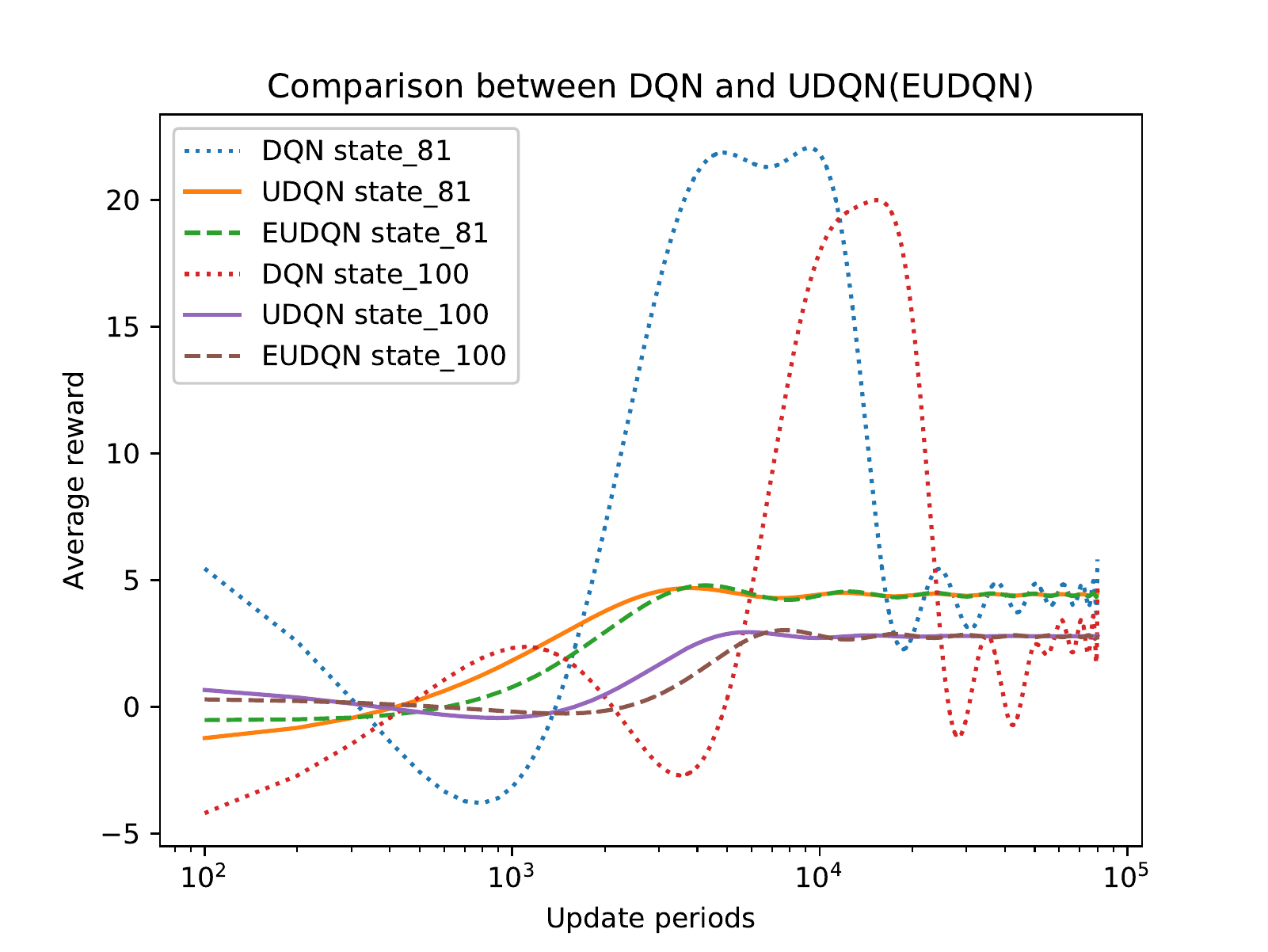}}
  \subfigure[]{
  \label{fig:UDQN-DQN-121-144}
  \includegraphics[width=0.49\textwidth]{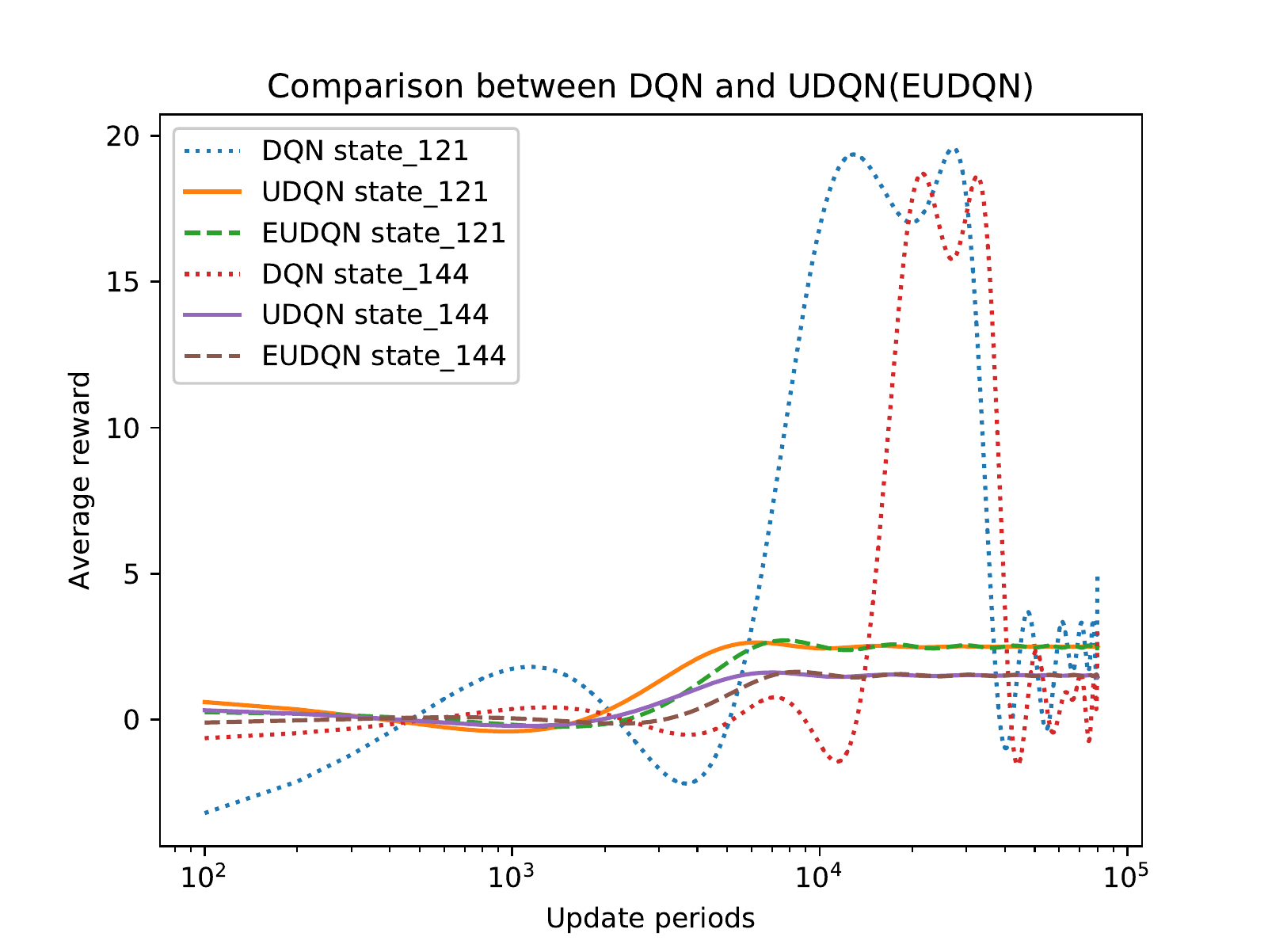}}
  \subfigure[]{
  \label{fig:UDQN-DQN-169-196}
  \includegraphics[width=0.49\textwidth]{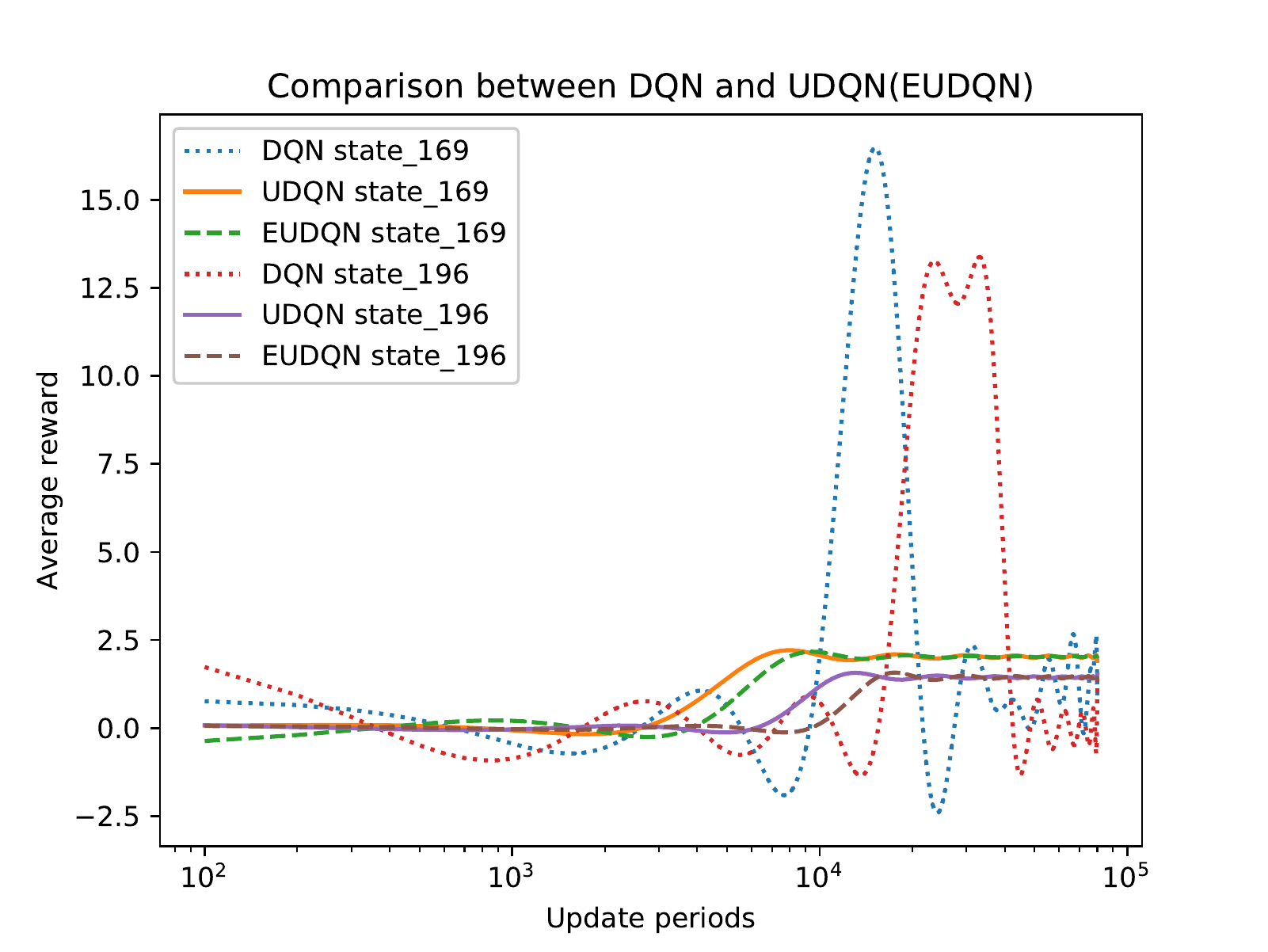}}
  \subfigure[]{
  \label{fig:UDQN-DQN-225-256}
  \includegraphics[width=0.49\textwidth]{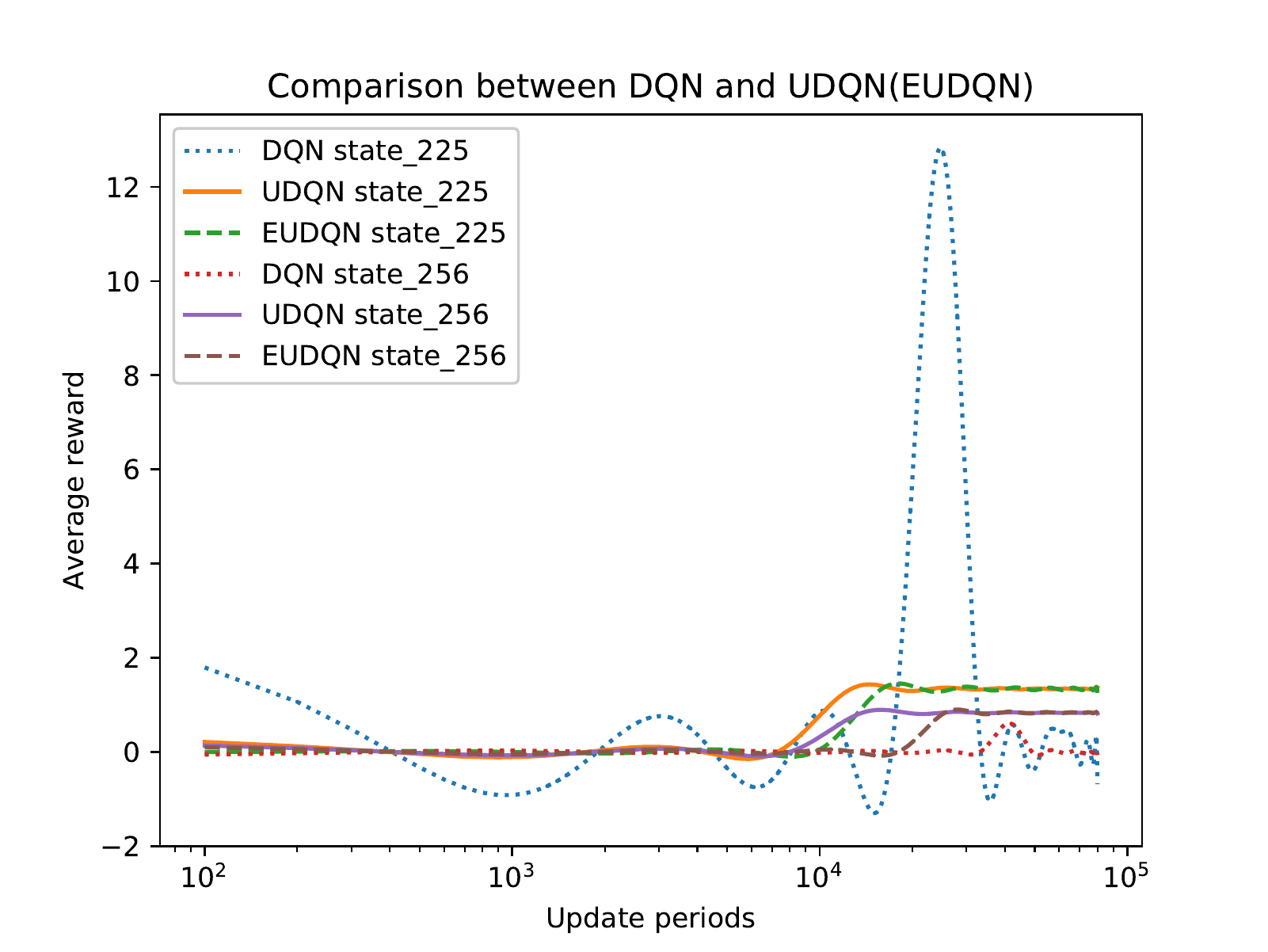}}
  \caption{\textbf{Computational efficiency comparison among DQN, UDQN and EUDN ranging from (a) 81-100 states in a square maze; (b) 121-144 states in a square maze; (c) 169-196 states in a square maze; (d) 225-256 states in a square maze.}}
  \label{fig:UDQN-DQN}
\end{figure}

% 数据图;  分析；
Then the results are shown in Figs.~\ref{fig:UDQN-DQN-81-100}-\ref{fig:UDQN-DQN-225-256}. These figures compare computational efficiency of UDQN and EUDQN with that of DQN ranging from $81$ to $256$ discrete states in a square maze. Specifically, an evaluation procedure is launched every $100$ update periods, which observes the agent starting from the start point and records the reward of each test episode. The timeout is set as $200$ steps, which means that one episode will terminate if the agent cannot arrive at the goal within $200$ steps. The average reward stands for the accumulated reward averaged over the number of steps consumed during each episode, and results of $100$ episodes are averaged for each evaluation procedure to ensure accuracy. In Fig.~\ref{fig:UDQN-DQN-81-100}, DQN, UDQN and EUDQN all converge within $80000$ update numbers due to relatively small state space, although DQN oscillates around the local optimal point with $50$-score bonus at the upper-left side of the goal (see Fig.~\ref{fig:Maze}) at the early stage. Also from Fig.~\ref{fig:UDQN-DQN-81-100}, DQN converges at around $41$ and $53$ thousand update numbers for $81$ and $100$ states, respectively, which are more than $10$ times of update periods necessary for the convergence of UDQN. From Fig.~\ref{fig:UDQN-DQN-121-144}, we can notice that DQN begins to lose its stability and diverges when faced with more obstacles, while UDQN and EUDQN converge much faster and more stably than DQN. When it comes to $169-256$ states given by Figs.~\ref{fig:UDQN-DQN-169-196} and \ref{fig:UDQN-DQN-225-256}, DQN diverges during $80000$ update periods while UDQN and EUDQN still keep robust. Although EUDQN is a bit slower than UDQN in convergence rate, it is much more sample-efficient considering the fact that number of samples consumed by UDQN is roughly $200$ times of that consumed by EUDQN given the hyperparameters of Table~\ref{Table:par_maze}. It is also noticed that the converged value decreases as the state space becomes larger, because the $150$-score bonus will be averaged over more steps from the start point to the goal. In conclusion, UDQN has the features of much higher computational efficiency and stability compared with DQN, and EUDQN does not lose much performance from UDQN.

% 说明一下各组参数下每次训练消耗的时间
We can also compare the convergence rate, in other words, the required time to converge, by estimating the average time cost for each update period. The time cost per update period is different even if the batch size of UDQN is the same as the mini-batch size of DQN, since DQN needs time to process memory storage and usage in experience replay. The ratio of average time cost of DQN to that of UDQN is around $1.73$ under the same device configuration. Multiplying the time cost per update period by the required update periods for the convergence of two algorithms (see Figs.~\ref{fig:UDQN-DQN-81-100}), we can judge that UDQN is far faster than DQN.

\begin{figure}
  \centering
  \subfigure[]{
  \label{fig:value-UDQN-81}
  \includegraphics[width=0.49\textwidth]{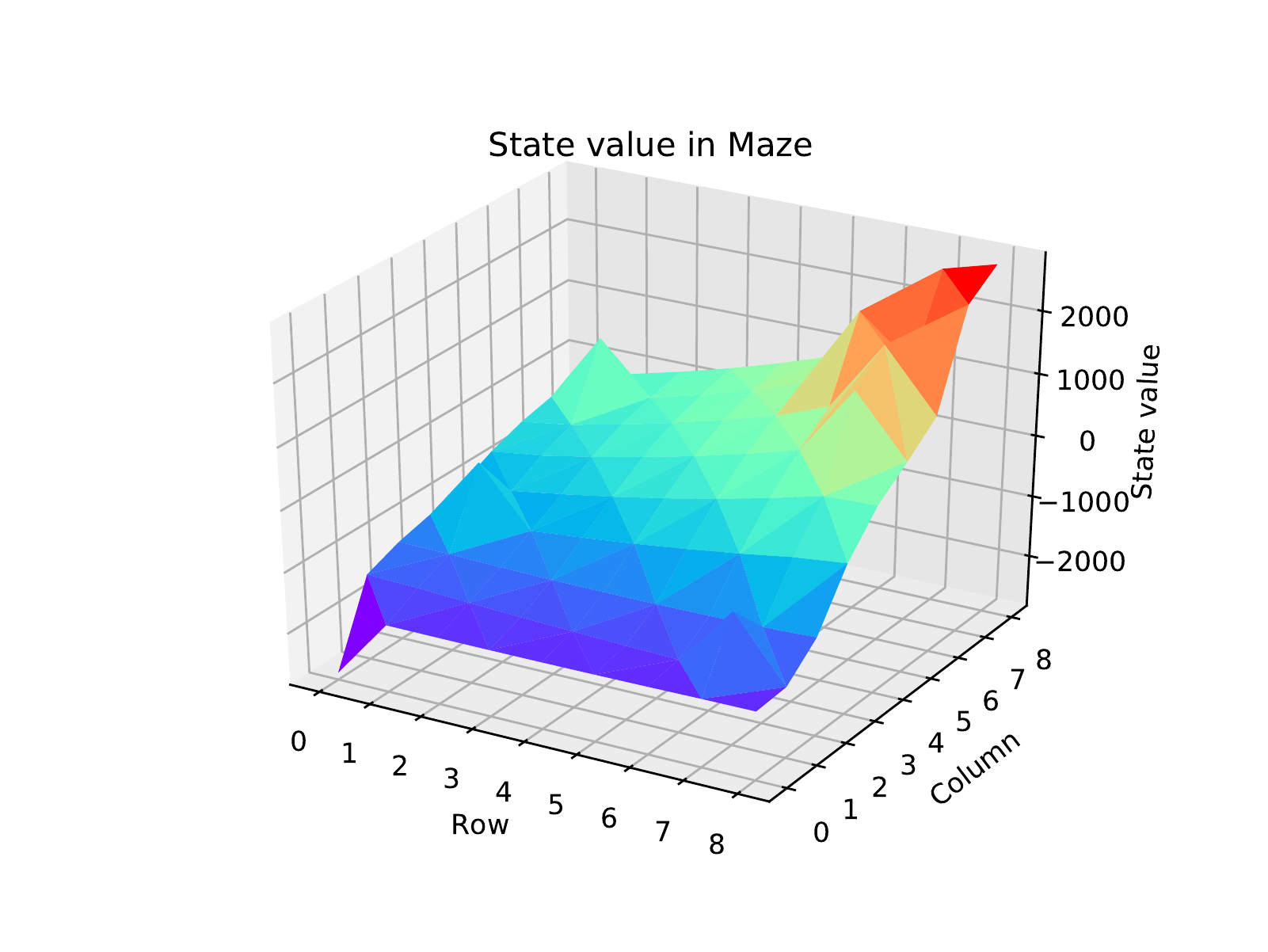}}
  \subfigure[]{
  \label{fig:value-UDQN-121}
  \includegraphics[width=0.49\textwidth]{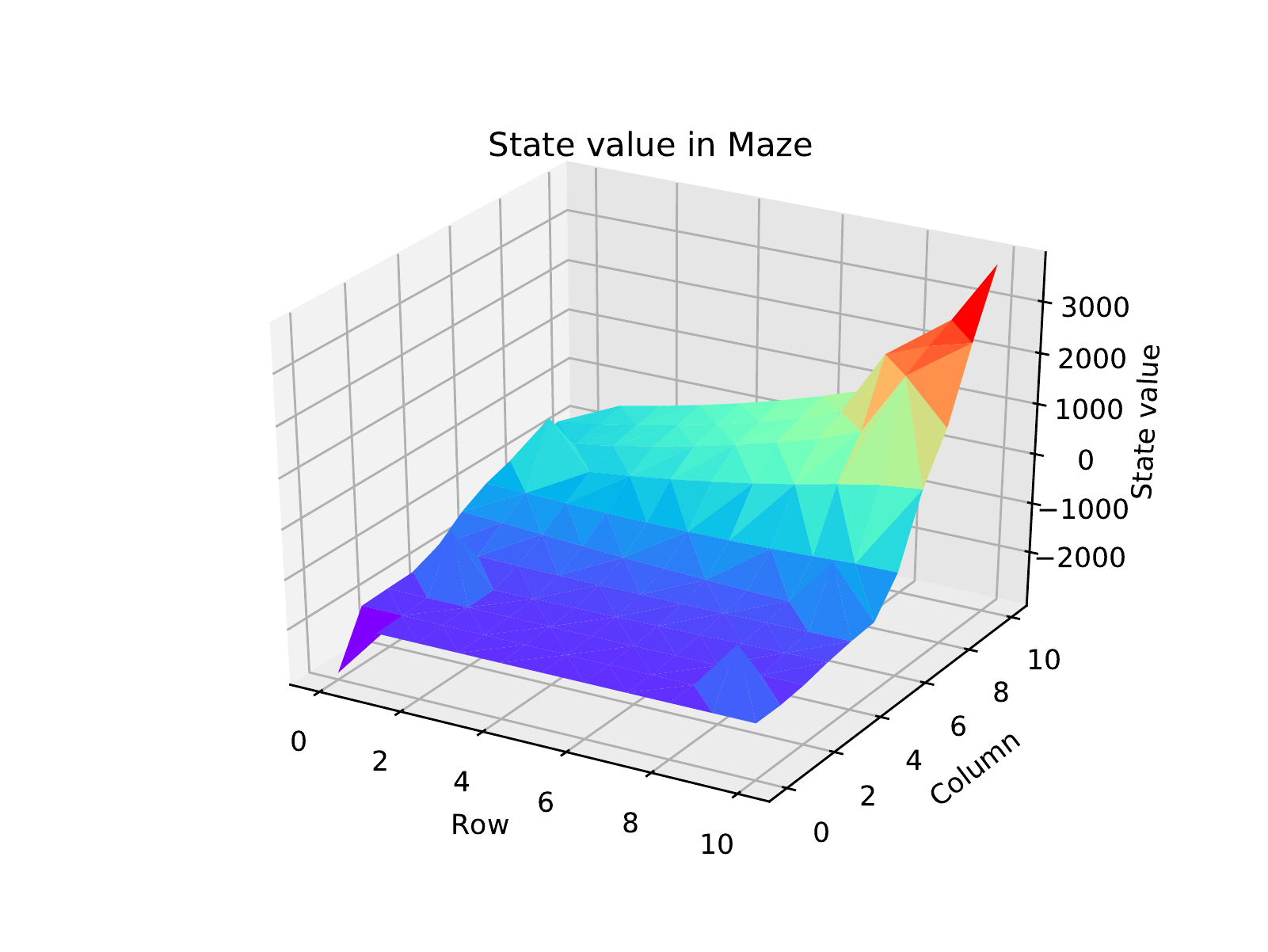}}
  \subfigure[]{
  \label{fig:value-UDQN-169}
  \includegraphics[width=0.49\textwidth]{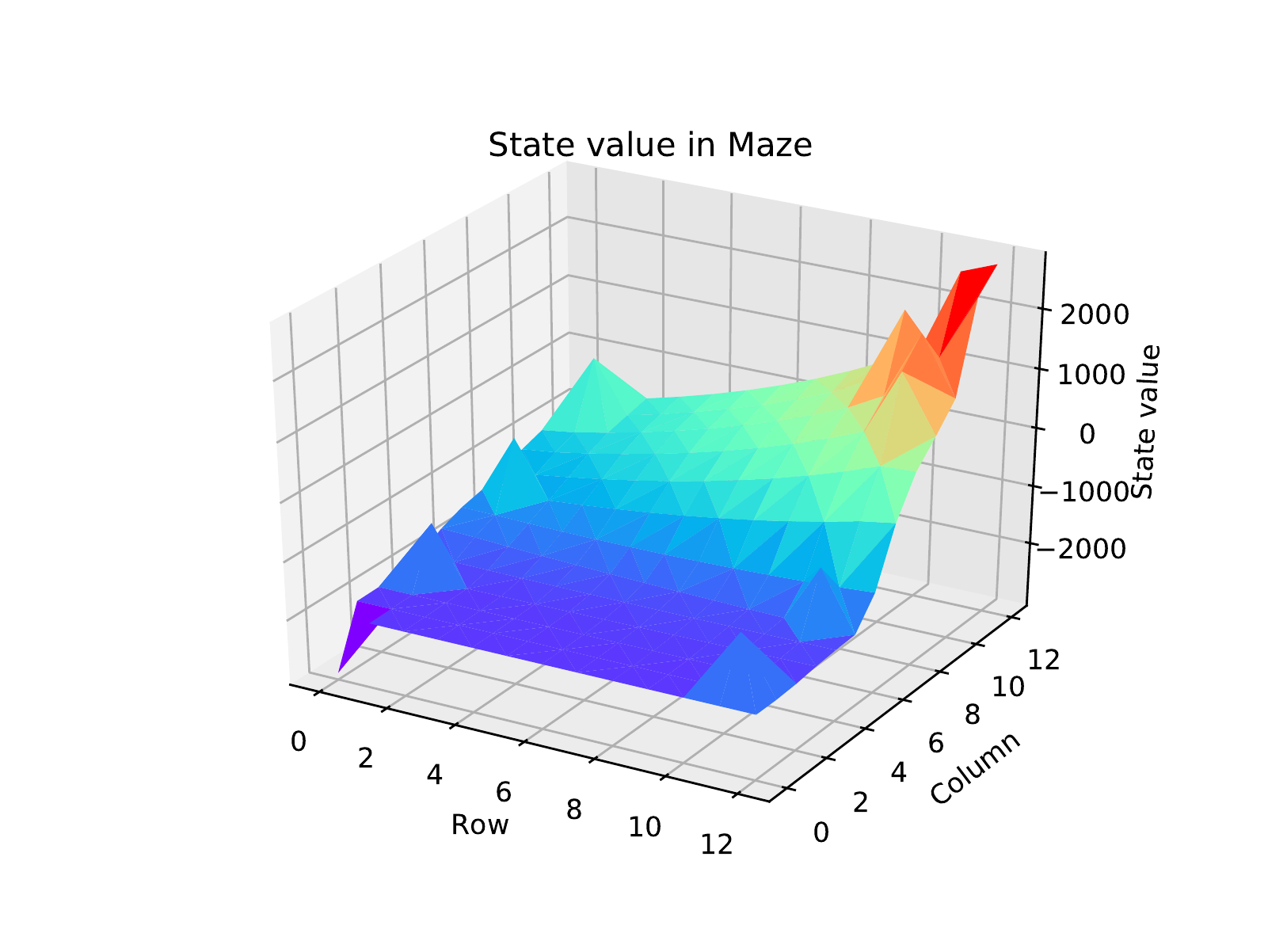}}
  \subfigure[]{
  \label{fig:value-UDQN-225}
  \includegraphics[width=0.49\textwidth]{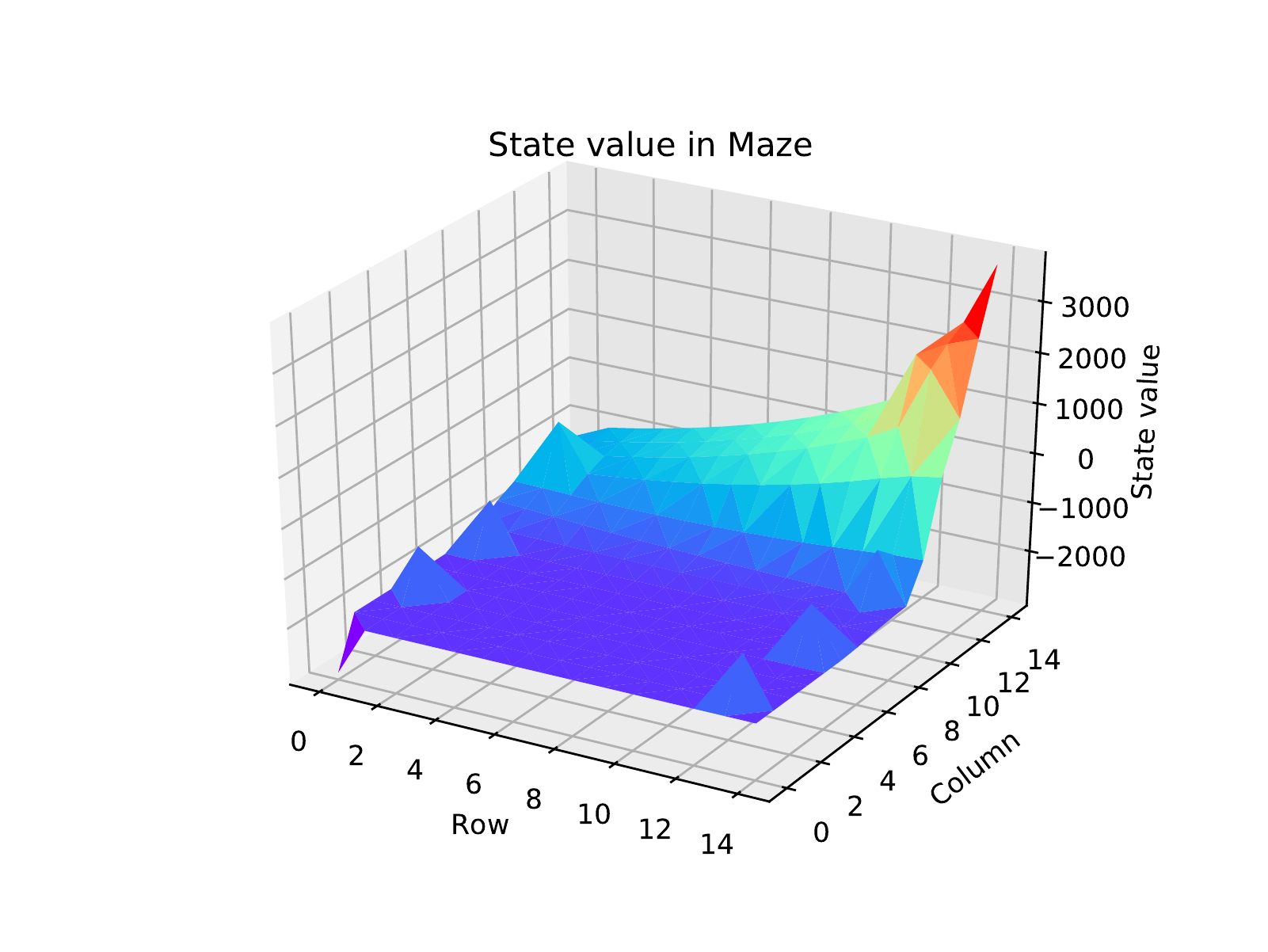}}
  \caption{\textbf{Value function of UDQN versus position in a square maze of (a) 81 states; (b) 121 states; (c) 169 states; (d) 225 states.}}
  \label{fig:value_pos}
\end{figure}

% 值函数相对于位置
Besides the computational efficiency, we plot the value function versus position of the agent in the maze environment in Fig.~\ref{fig:value_pos} to observe how the optimal function learnt by the UDQN algorithm is distributed over the state space. From Fig.~\ref{fig:Maze}, we find nearly all areas are surrounded by obstacles. This is why most areas in Fig.~\ref{fig:value_pos} are flat with low values. Since we make the walls have the same minus feedback as that of obstacles, the initial state just has one choice to avoid minus reward, thus having the lowest value. We notice that the area near the goal has a relative high value, and the peak of value function is located exactly at the goal, which is reasonable because the goal rewards the highest score. Besides, it is noticeable that there exists an area higher than its surroundings aside the goal in all the four scenarios from Figs.~\ref{fig:value-UDQN-81}-\ref{fig:value-UDQN-225}. This special area can date back to the upper-left side of the goal which has a $50$-score bonus. Moreover, it is interesting to observe several local peaks in Figs.~\ref{fig:value-UDQN-81}-\ref{fig:value-UDQN-225}. Each of these local peaks corresponds to a cross point, the point which has three outlets, i.e., three adjacent locations giving nonnegative rewards. Take the maze with $81$ states having $9$ rows and $9$ columns as an example, there are four cross points according to Fig.~\ref{fig:Maze}, whose coordinates are respectively $(1,7)$, $(3,1)$, $(5,7)$ and $(7,1)$, thus causing four local peaks in Fig.~\ref{fig:value-UDQN-81}. Accordingly, every two rows will bring one more cross point. Therefore, we see six local peaks in Fig.~\ref{fig:value-UDQN-169}, which have six cross points. As for mazes of $121$ and $225$ states in Figs.~\ref{fig:value-UDQN-121} and \ref{fig:value-UDQN-225}, there should have been respectively five and seven local peaks, but unluckily one of them is occupied by the $50$-score subgoal.

\subsection{Robot arm}

% Robot arm 环境
Due to the trouble caused by redefinition of environment, we take the "robot arm" experiment, a grasp and move task, for example. The environment of "robot arm" experiment is shown in Fig.~\ref{fig:Robot_arm}. In this figure, the green circle at the center of the space is fixed to confine one side of the arm. Other circles are used to connect these limbs so that the finger represented by the end of the arm can travel all over the whole space. We just plot three sections to stand for a general arm that contains any potential number of sections. In the experiment setting, the goal represented by the yellow box is randomly mobile, so the state representation includes both the positions of joints and their relative positions to the goal. Accordingly, the action space has the same dimension as the number of sections, containing angles for the sections to rotate by. The rewards are set as the minus distance from the finger to the goal plus a bonus, which is $1$ when the finger is located within the domain of goal box. Once the finger catches the goal, i.e., the position of the finger locates within the goal box, it needs to hold on to the mobile goal for several steps to ensure stability of the grasp task. Otherwise, the samples which fall out of the goal or have finished the grasp steps are reset to prepare for another iteration. Although Algorithm \ref{code:UDDPG} provides a general idea for UDDPG programming, actually several steps need to be added or revised when applied to the grasp and move task. Specifically, there should be job-done signals after executing actions to determine whether to continue holding on to the goal or start a new iteration.
%UDDPG algorithm for grasp and move task is given by Algorithm \ref{code:UDDPG_robot_arm}.
%
%\begin{algorithm}[h]
%\caption{UDDPG Algorithm for Grasp and Move Task}
%\label{code:UDDPG_robot_arm}
%\begin{algorithmic}[1]
%\State $\mathbf{Input}$: The batch size $N$, the batch maximum $M$, and the soft update parameter $\tau$.
%\State $\mathbf{Initialization}$: Initialize the network parameters $(\omega,\theta,\omega',\theta')\gets(\omega_0,\theta_0,\omega_0',\theta_0')$ arbitrarily.
%\For{$i=1,M$}
%\State Uniformly sample $S_i=(s_{i,1},s_{i,2},\cdots,s_{i,N})|_{n:h_{i,n}==0}$ over the whole state space;
%\State Choose actions $A_i=(a_{i,1},a_{i,2},\cdots,a_{i,N})$ for $S_i$ according to the current actor network $\mu(S_i|\theta_i)$ added by an exploration noise;
%\State Execute actions $A_i$, get next states $S'_i=(s'_{i,1},s'_{i,2},\cdots,s'_{i,N})$, immediate rewards $R_i=(r_{i,1},r_{i,2},\cdots,r_{i,N})$, and job-done signs $H_i=(h_{i,1},h_{i,2},\cdots,h_{i,N})$;
%\State Maximize the expected return shown in \eqref{eq:a_appro} by gradient ascent, and then update $\theta_i$;
%\State Minimize the loss function shown in \eqref{eq:loss_DDPG} by gradient decent, and then update $\omega_i$;
%\State Execute the "soft" target updates shown in \eqref{eq:soft_update} to update $\theta_{i+1}'$ and $\omega_i'$;
%\State $S_i|_{n:h_{i,n}==1}\gets S'_i|_{n:h_{i,n}==1}$;
%\EndFor
%\end{algorithmic}
%\end{algorithm}

% 参数与表格 hyperparameter values
We implement DDPG and UDDPG (EUDDPG) algorithms on the "robot arm" environment for a fair comparison, using the same hyperparameter values listed in Table~\ref{Table:par_robot}. Both DDPG and UDDPG (UDDPG) exploit the prior knowledge of space area in Fig.~\ref{fig:Robot_arm} to ensure convergency.

\begin{figure}
  \centering
  \subfigure[]{
  \label{fig:UDDPG-DDPG-2-3}
  \includegraphics[width=0.49\textwidth]{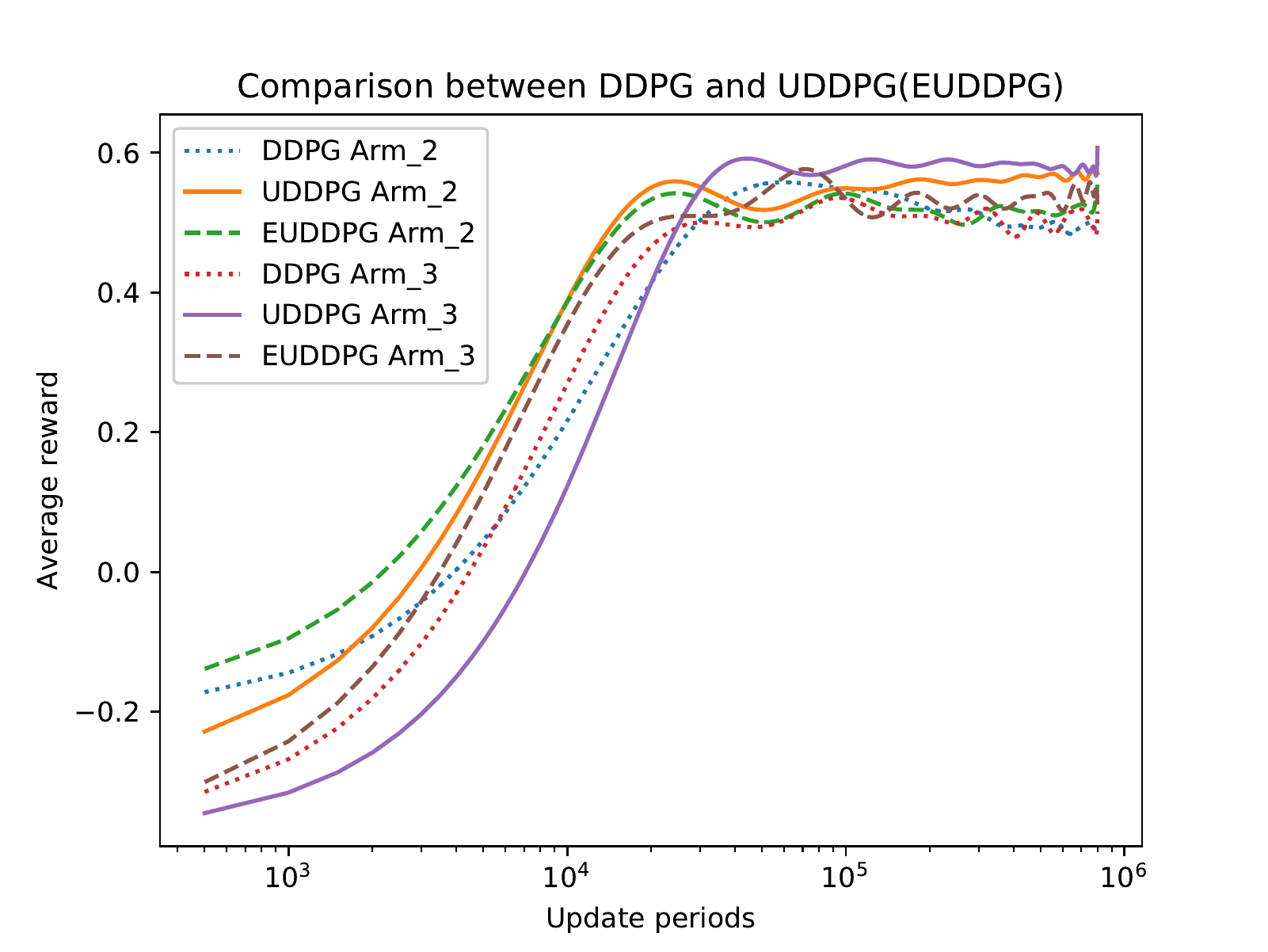}}
  \subfigure[]{
  \label{fig:UDDPG-DDPG-4-5}
  \includegraphics[width=0.49\textwidth]{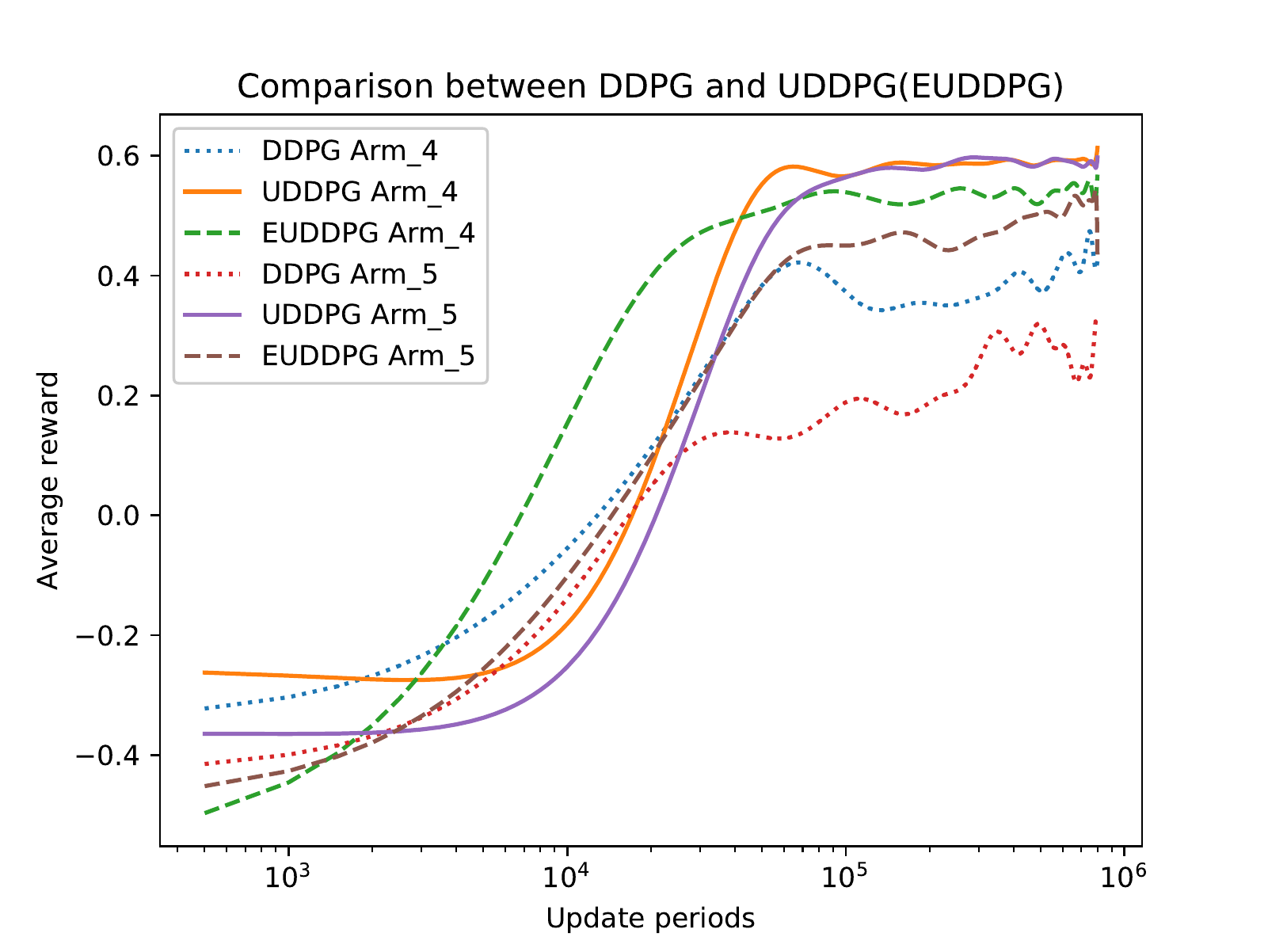}}
  \subfigure[]{
  \label{fig:UDDPG-DDPG-6-7}
  \includegraphics[width=0.49\textwidth]{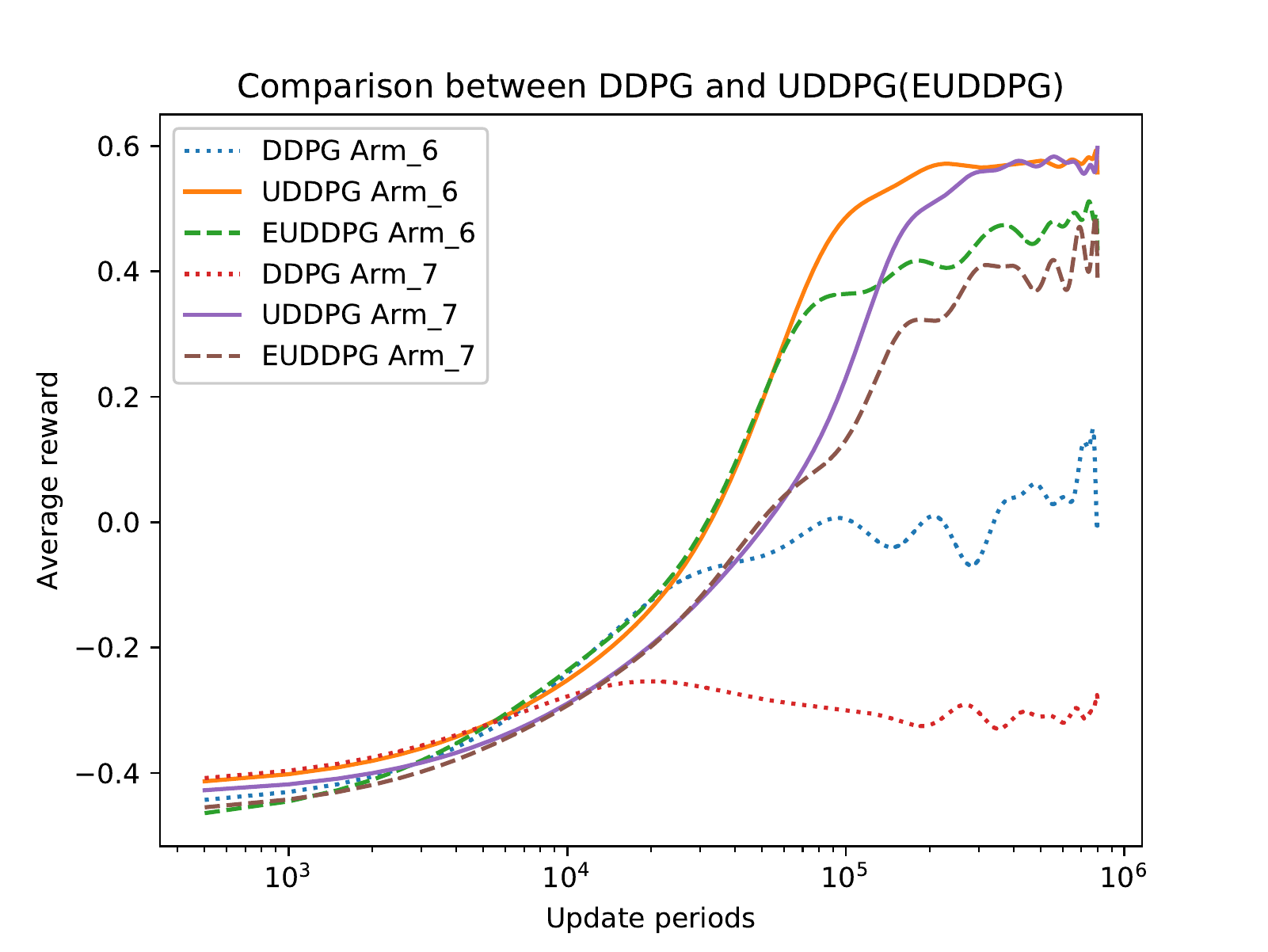}}
  \subfigure[]{
  \label{fig:UDDPG-DDPG-8-9}
  \includegraphics[width=0.49\textwidth]{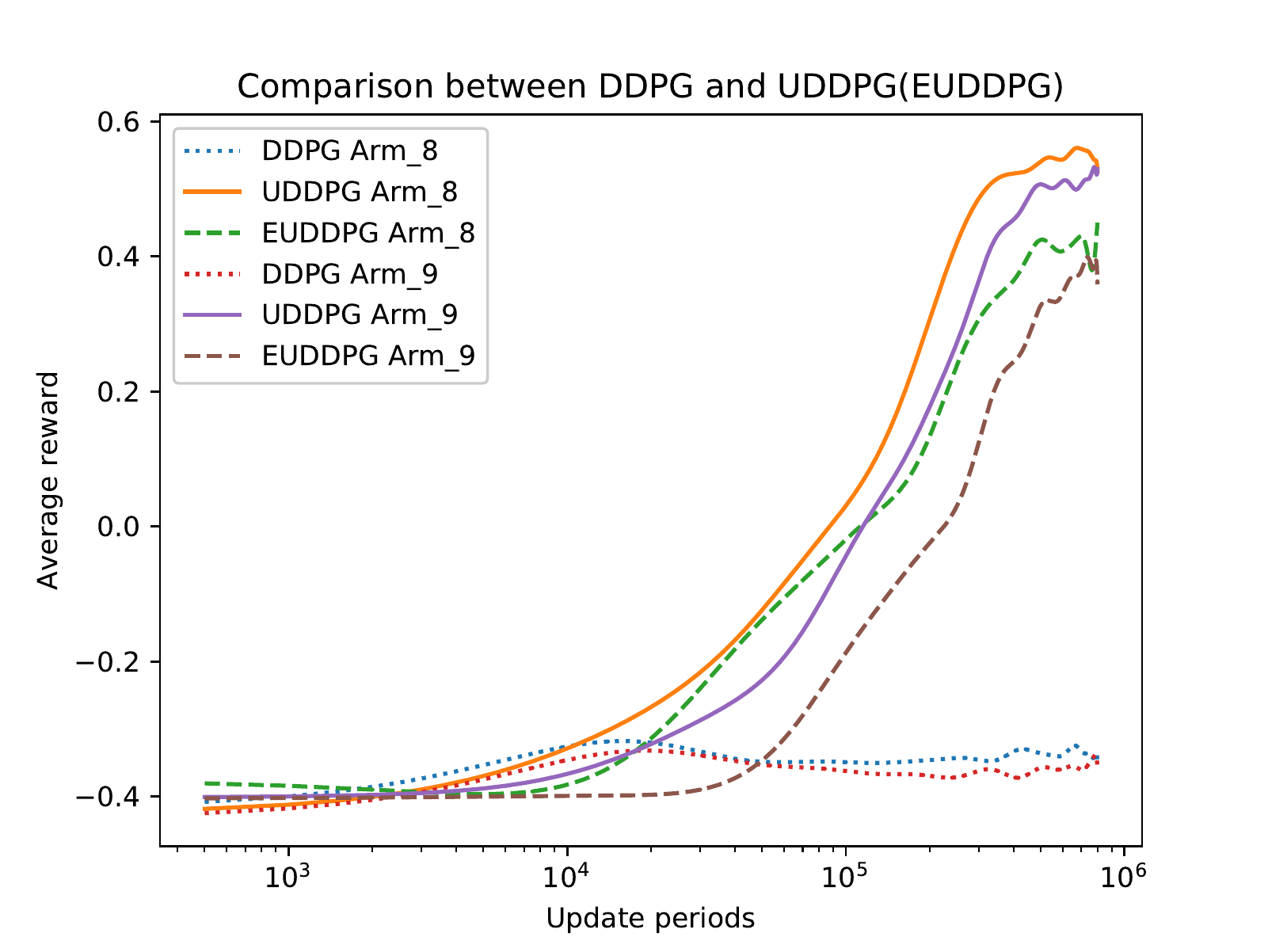}}
  \caption{\textbf{Computational efficiency comparison between DDPG and UDDPG for a robot arm with (a) 2-3 sections; (b) 4-5 sections; (c) 6-7 sections; (d) 8-9 sections.}}
  \label{fig:UDDPG-DDPG}
\end{figure}

% 数据图;  分析；
For generality, we gradually increase the number of sections in the "robot arm" experiment to observe the performance of algorithms faced with growing state dimension. Notably, one more section will increase the state dimension by $4$, including the coordinates ($2$ dimensions) of joints and their relative coordinates to the goal. Then Fig.~\ref{fig:UDDPG-DDPG} compares computational efficiency of UDDPG and EUDDPG with that of DDPG ranging from $2-9$ sections ($9-37$ state dimensions) by fitting scatterplots generated from $800$ thousand update periods. Specifically, the evaluation interval is $500$ update periods, which observes the robot arm initialized randomly and moving its finger gradually closer to the randomly located goal, until it completes the grasp task or sees the timeout. The timeout is set as $100$ steps, which means that one episode will restart if the robot arm cannot fulfill the grasp task within $100$ steps. During each episode, the accumulated reward will be recorded and averaged over the number of consumed steps, and the final result on the Y-axis of Fig.~\ref{fig:UDDPG-DDPG} is the average reward of $100$ episodes for each evaluation procedure.
In Fig.~\ref{fig:UDDPG-DDPG-2-3}, when the state dimension is relatively low, it is noticeable that all DDPG, UDDPG and EUDDPG can converge to a level within $800$ thousand update periods. From Fig.~\ref{fig:UDDPG-DDPG-4-5}, we can see that as state dimension grows, the convergence of DDPG becomes unstable and low-level while UDDPG and EUDDPG remain robust although EUDDPG is a bit lower than UDDPG in convergence rate and level. Due to the mobility of the goal in the grasp and move task, different convergence levels can be interpreted as the sensitivity of algorithms that enables the robot arm to follow the behavior of the goal, i.e., higher convergence level shows the agent can react more promptly to the randomly mobile goal. In common sense, larger state dimension will render slower convergence speed, lower convergence level (less sensitivity) and higher instability, which is conspicuous from the observation of DDPG curves in Figs.~\ref{fig:UDDPG-DDPG-6-7} and \ref{fig:UDDPG-DDPG-8-9}. However, UDDPG proves to be robust in all the three above-mentioned aspects. EUDDPG is slightly inferior to UDDPG in these aspects, but more sample-efficient.

\section{Conclusion}  \label{Conclusion}
% 本文
This paper aims to develop a general framework to embrace existing and future RL algorithms. Theoretical analysis first characterizes UDRL (EUDRL) as the unbiased approximation for CPU or GPU computation and has the strength of uniform convergence. Then experimental results show that our UDRL framework is more computationally efficient and stable than DRL for both discrete and continuous state-action spaces. Although EUDRL is a bit inferior to UDRL in convergence rate, it is more sample-efficient from the perspective of sample utilization. Overall, the UDRL (EUDRL) can replace the existing DRL based on theoretical analysis and experimental validation.

\begin{table}
\centering
\caption{\textbf{List of hyperparameters of the discrete maze experiment}}
\begin{tabular}{c c c}
\hline
Hyperparameter & Value & Description \\
\hline
learning rate  & 0.001 & The learning rate used by gradient descent optimizer\\
initial exploration rate & 0.9 & Initial value of $\epsilon$ in $\epsilon$-greedy policy\\
final exploration rate & 0.0001 & Minimun value of $\epsilon$ in $\epsilon$-greedy policy\\
discount factor & 0.9 & The discount horizon factor $\gamma$ to estimate the target value\\
observation size & 2500 & The number of samples collected before training in DQN\\
batch size & 200 & The sample size of UDQN per batch\\
mini-batch size & 200 & The sample size of DQN and EUDQN from replay memory per step and per batch\\
mini-batch maximum & 200 & The number of EUDQN mini-batch samples per batch\\
replay memory size & 10000 & The size of replay buffer used in DQN and EUDQN\\
\hline
\end{tabular}
\label{Table:par_maze}
\end{table}

\begin{table}
\centering
\caption{\textbf{List of hyperparameters of the "robot arm" experiment}}
\begin{tabular}{c c c}
\hline
Hyperparameter & Value & Description \\
\hline
learning rate & 0.001 & The learning rate used by gradient descent optimizer\\
initial exploration variance & 1.0 & Initial variance of gaussian exploration noise added to actions\\
variance decay rate & 0.9995 & The exploration variance multiplied by the rate per training\\
soft update parameter  & 0.01 & $\tau$ used in "soft" target updates\\
discount factor & 0.9 & The discount horizon factor $\gamma$ to estimate the target value\\
maximum episode steps & 200 & The timeout steps for DDPG\\
batch size & 200 & The sample size of UDDPG per batch\\
mini-batch size & 200 & The sample size of DDPG and EUDDPG from replay memory per step and per batch\\
mini-batch maximum & 200 & The number of EUDDPG mini-batch samples per batch\\
replay memory size & 30000 & The size of replay buffer used in DDPG and EUDDPG\\
\hline
\end{tabular}
\label{Table:par_robot}
\end{table}

\section{Appendix}  \label{Appendix}
\subsection{Proof of Lemma~\ref{seq_iid}}
\begin{pf}
Assume each of $\{s_n\}_{n=1,\cdots,N}$ shares the same probability $p_s(s)$, then
\begin{align}
&p_a(a_n)=\int_{s_n}p_s(s_n)\pi(a_n|s_n)d s_n=\int_s p_s(s)\pi(a_n|s)ds=\int_s p(s,a)ds,\nonumber \\
&p_r(r_n)=\int_{s_n}\int_{a_n}p(r_n|s_n,a_n)p(s_n,a_n)d s_n d a_n=\int_s\int_a p(r_n|s,a)p(s,a)ds da,\nonumber\\
&p_{s'}(s'_n)=\int_{s_n}\int_{a_n}p(s'_n|s_n,a_n)p(s_n,a_n)d s_n d a_n=\int_s\int_a p(s'_n|s,a)p(s,a)dsda, \label{eq:p_slot}
\end{align}
where $\pi(\cdot|s)$ is the behavior policy, $p(s,a)$ stands for the joint distribution of $(s,a)$, and $p(\cdot|s,a)$ is the transition probability of rewards or next states. Specifically, if the transitions are deterministic, then $p_r(\cdot)$ and $p_{s'}(\cdot)$ are both equal to $1$. Overall, we can see from \eqref{eq:p_slot} that $\{a_n\}_{n=1,\cdots,N}$, $\{r_n\}_{n=1,\cdots,N}$ and $\{s'_n\}_{n=1,\cdots,N}$ are IID, respectively.
\end{pf}

\subsection{Proof of Theorem~\ref{evalu_UDRL}}
\begin{pf}
First, using the Bellman equation as an iterative update, i.e.,
\begin{align}
Q(s,a|\omega)=r(s,a)+\gamma\mathbb{E}_{s'\sim p(s'|s,a)}\left[Q(s',\mu(s')|\omega')\right],\forall (s,a)\in\chi\times A,\label{eq:Bell_eq}
\end{align}
where $p(s'|s,a)$ is the transition probability of the next state $s'$ given the current state-action pair $(s,a)$, and $\chi\times A$ is the state-action space, then the action-value function will finally converge to its optimal value.

Second, \ref{eq:loss_DRL} describes a general model-free stochastic approximation algorithm. According to \cite{szepesvari2010algorithms}, if the learning rate satisfies the Robbins-Monro (RM) conditions, i.e.,
\begin{align}
\sum_{t=0}^{\infty}\alpha_t=\infty,\quad \sum_{t=0}^{\infty}\alpha_t^2<+\infty ,\label{eq:RM_con}
\end{align}
the stochastic approximation $Q(s,a|\omega)\gets r+\gamma Q(s',\mu(s')|\omega')$ will converge to \eqref{eq:Bell_eq}, and even the commonly used constant learning rate can make algorithms converge in distribution. Besides, within the scope of current environment DRL is dealing with, transition probabilities are normally deterministic, which means that the next state $s'$ is unique given a specific state-action pair $(s,a)$. In that case, the stochastic approximation is exactly equal to \eqref{eq:Bell_eq}.

Third, to update the network parameters, the loss function should be the expectation of stochastic approximation errors over the whole state space, given by \eqref{eq:loss_DRL}. If the current states are IID, then \eqref{eq:loss_UDRL} is the unbiased approximation of \eqref{eq:loss_DRL}, which means that \eqref{eq:loss_UDRL} will uniformly converge to \eqref{eq:loss_DRL} given enough samples.

Overall, the network parameters will finally converge to their optimal values by minimizing \eqref{eq:loss_UDRL}.

\end{pf}

\subsection{Proof of Theorem~\ref{impro_UDRL}}
\begin{pf}
According to the iterative update of RL algorithms, we have
\begin{align}
&\mathbb{E}_{(s,a)\sim\bm{P}}\left[Q(s,a|\omega_{i+1})\right]=\mathbb{E}_{(s,a,r,s')\sim\bm{P}}\left[r(s,a)+\gamma Q(s',\mu_i(s')|\omega')\right]\\
&\approx\mathbb{E}_{(s,a,r,s')\sim\bm{P}}\left[r(s,a)\right]+\frac{\gamma}{N}\sum_{n=1}^N Q(s_n',\mu_i(s_n')|\omega')\\
&\geq\mathbb{E}_{(s,a,r,s')\sim\bm{P}}\left[r(s,a)\right]+\frac{\gamma}{N}\sum_{n=1}^N Q(s_n',\mu_{i-1}(s_n')|\omega')\\
&\approx\mathbb{E}_{(s,a,r,s')\sim\bm{P}}\left[r(s,a)+\gamma Q(s',\mu_{i-1}(s')|\omega')\right]=\mathbb{E}_{(s,a)\sim\bm{P}}\left[Q(s,a|\omega_i)\right]
, \label{eq:iter_update}
\end{align}
\end{pf}

\end{document}